\documentclass[10pt,twocolumn,letterpaper]{article}
\pdfoutput=1
\usepackage[pagenumbers]{cvpr} %
\usepackage[accsupp]{axessibility} %
\usepackage[dvipsnames]{xcolor}

\newcommand{\name}{{Video2Game}}
\newcommand{\todocite}[1]{\textcolor{blue}{Citation needed []}}
\newcommand{\zhihao}[1]{{\color{black}{{#1}}}}
\newcommand{\hongchi}[1]{\textcolor{black}{#1}}

\makeatletter
\def\@onedot{\ifx\@let@token.\else.\null\fi\xspace}
\DeclareRobustCommand\onedot{\futurelet\@let@token\@onedot}

\def\eg{\emph{e.g}\onedot}

\newcommand\shortcite[2][]{%
  \ifNAT@numbers\cite[#1]{#2}\else\citeyearpar[#1]{#2}\fi}
\newcommand{\ignorethis}[1]{}

\makeatletter
\DeclareRobustCommand\onedot{\futurelet\@let@token\@onedot}
\def\@onedot{\ifx\@let@token.\else.\null\fi\xspace}

\def\eg{\emph{e.g}\onedot}

\makeatother

\definecolor{citecolor}{RGB}{34,139,34}
\definecolor{mydarkblue}{rgb}{0,0.08,1}
\definecolor{mydarkgreen}{rgb}{0.02,0.6,0.02}
\definecolor{mydarkred}{rgb}{0.8,0.02,0.02}
\definecolor{mydarkorange}{rgb}{0.40,0.2,0.02}
\definecolor{mypurple}{RGB}{111,0,255}
\definecolor{myred}{rgb}{1.0,0.0,0.0}
\definecolor{mygold}{rgb}{0.75,0.6,0.12}
\definecolor{myblue}{rgb}{0,0.2,0.8}
\definecolor{mydarkgray}{rgb}{0.66,0.66,0.66}

\newcommand{\bbR}{{\mathbb{R}}}

\newcommand{\bx}{\mathbf{x}}

\newcommand{\bS}{\mathbf{S}}
\newcommand{\bs}{\mathbf{s}}

\newcommand{\bw}{\mathbf{w}}

\newcommand{\bV}{\mathbf{V}}

\newcommand{\bn}{\mathbf{n}}

\newcommand{\bB}{\mathbf{B}}
\newcommand{\bD}{\mathbf{D}}

\newcommand{\bC}{\mathbf{C}}

\newcommand{\bF}{\mathbf{F}}
\newcommand{\bo}{\mathbf{o}}
\newcommand{\bc}{\mathbf{c}}
\newcommand{\bT}{\mathbf{T}}

\newcommand{\bd}{\mathbf{d}}
\newcommand{\br}{\mathbf{r}}

\usepackage{bbding} 
\definecolor{cvprblue}{rgb}{0.21,0.49,0.74}
\definecolor{darkgreen}{rgb}{0,0.5,0}
\usepackage[pagebackref,breaklinks,colorlinks,citecolor=cvprblue]{hyperref}
\usepackage{multirow}
\usepackage{bm}
\usepackage{amsmath}
\usepackage{amssymb}
\usepackage{adjustbox}
\usepackage{makecell}
\usepackage{xspace}
\usepackage{pifont}
\newcommand{\cmark}{\textcolor{darkgreen}{\ding{51}}}
\newcommand{\xmark}{\textcolor{red}{\ding{55}}}%
\usepackage[ruled]{algorithm2e}
\def\confName{CVPR}
\def\confYear{2024}

\title{Video2Game: Real-time, Interactive, Realistic and Browser-Compatible Environment from a Single Video}

\author{Hongchi Xia$^{1,2}$
\quad
Zhi-Hao Lin$^{1}$
\quad 
Wei-Chiu Ma$^{3}$
\quad 
Shenlong Wang$^{1}$\\
$^{1}$University of Illinois Urbana-Champaign
\quad
$^{2}$Shanghai Jiao Tong University
\quad 
$^{3}$Cornell University\\
\large{\href{https://video2game.github.io/}{https://video2game.github.io/}}
}

\begin{document}

\twocolumn[{%
\renewcommand\twocolumn[1][]{#1}%
\maketitle
\vspace{-10mm}
\captionsetup{type=figure}
\begin{center}
\includegraphics[width=\textwidth]{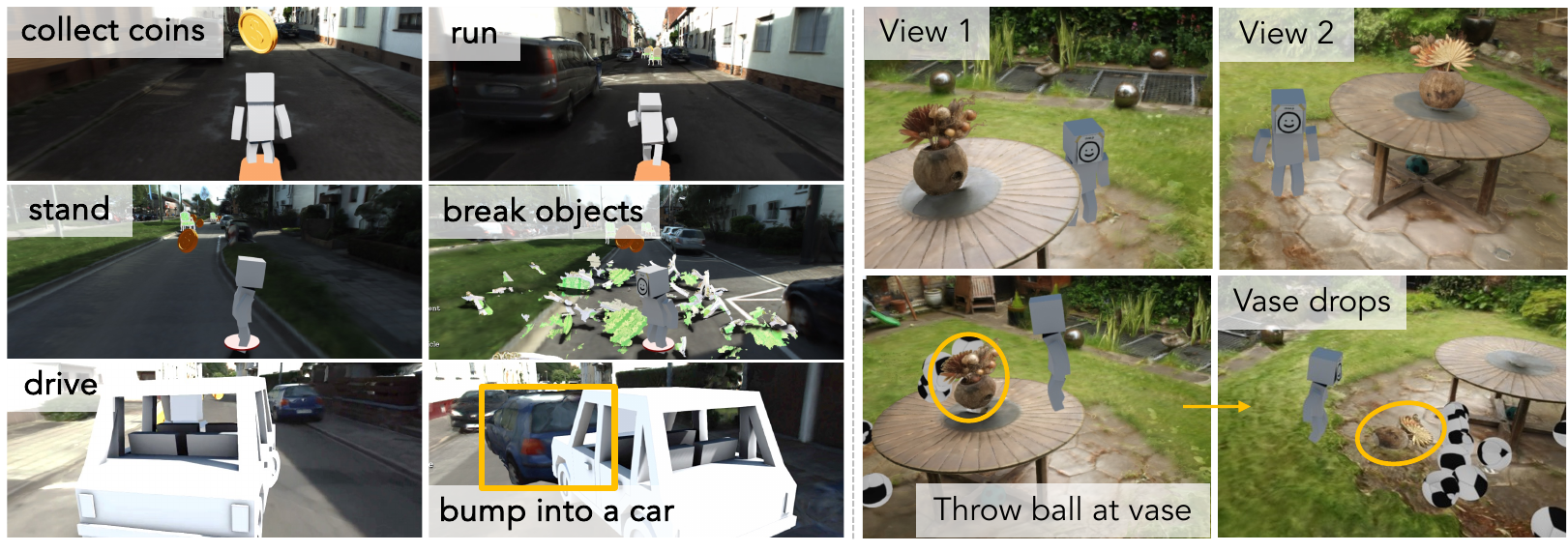}
\end{center}
\vspace{-6mm}
\captionof{figure}{{\bf Video2Game} {takes an input video of an arbitrary scene and automatically transforms it into a real-time, interactive, realistic and browser-compatible environment. 
The users can freely explore the environment and interact with the objects in the scene. 
} }
\label{fig:teaser}
}
\vspace{5mm}
]

\begin{abstract}
Creating high-quality and interactive virtual environments, such as games and simulators, often involves complex and costly manual modeling processes. 
In this paper, we present Video2Game, a novel approach that automatically converts videos of real-world scenes into realistic and interactive game environments. 
At the heart of our system are three core components:
(i) a neural radiance fields (NeRF) module that effectively captures the geometry and visual appearance of the scene;
(ii) a mesh module that distills the knowledge from NeRF for faster rendering;
and (iii) a physics module that models the interactions and physical dynamics among the objects.
By following the carefully designed pipeline, one can construct an interactable and actionable digital replica of the real world. 
We benchmark our system on both indoor and large-scale outdoor scenes. 
We show that we can not only produce highly-realistic renderings in real-time, but also build interactive games on top. 
\end{abstract}
    
    \vspace{-20px}
\section{Introduction}

Crafting a visually compelling and interactive environment is crucial for immersive experiences in various domains, such as video games, virtual reality applications, and self-driving simulators.
This process, however, is complex and expensive.
It demands the skills of experts in the field and the use of professional software development tools~\cite{haas2014history,unrealengine}.
For instance, Grand Theft Auto V~\cite{gta5}, known for its intricately detailed environment, was one of the most expensive video games ever developed, with a budget over \$265 million primarily for asset creation. 
Similarly, the development of the CARLA autonomous driving simulator~\cite{dosovitskiy2017carla} involves a multidisciplinary team of 3D artists, programmers, and engineers to meticulously craft and texture the virtual cityscapes, creating its lifelike environments.

\begin{figure*}[t]
    \centering
    \includegraphics[width=\textwidth]{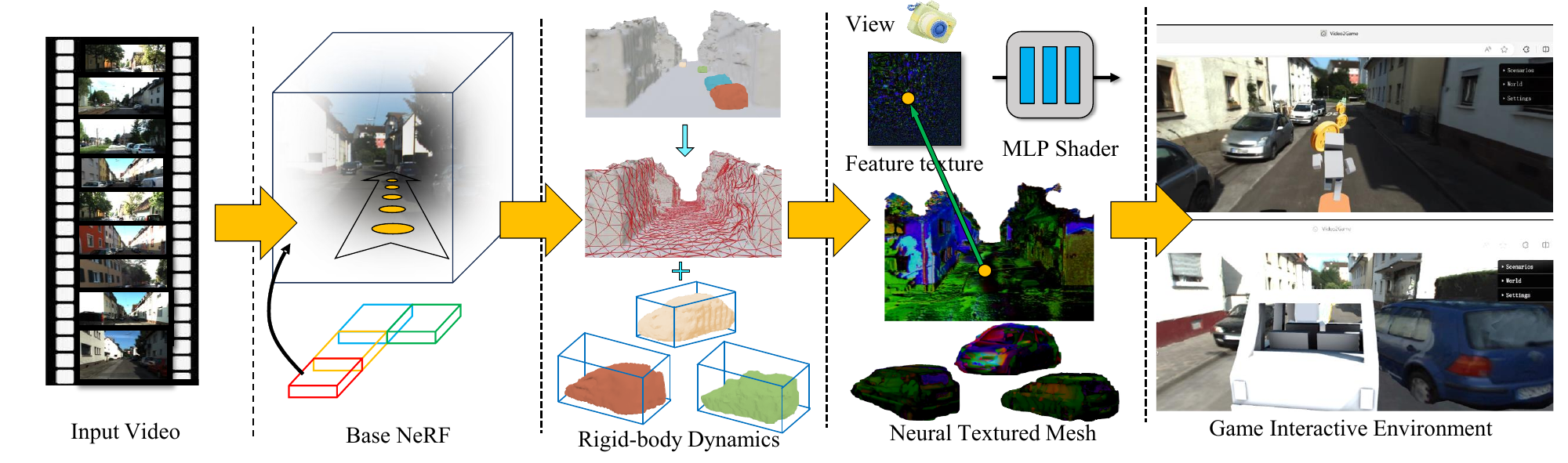}
    \vspace{-6mm}
    \caption{\textbf{Overview of Video2Game:}  
    Given multiple posed images from a single video as input, we first construct a large-scale NeRF model that is realistic and possesses high-quality surface geometry. We then transform this NeRF model into a mesh representation with corresponding rigid-body dynamics to enable interactions. We utilize UV-mapped neural texture, which is both expressive and compatible with game engines. Finally, we obtain an interactive virtual environment that virtual actors can interact with, can respond to user control, and deliver high-resolution rendering from novel camera perspectives -- all in real-time.
    }
    \label{fig:overview}
\end{figure*}

An appealing alternative to extensive manual modelling is creating environments directly from the real world. For instance, photogrammetry, a technique for constructing digital replicas of objects or scenes from overlapping real-world photographs, has been utilized for environment creation~\cite{poullis2004photogrammetric,portales2010augmented}. Success stories also span various games and simulators. However, most use cases are limited to creating object assets and necessitate significant post-processing, such as material creation, texturing, and geometry fixes~\cite{thomas2019study}. 
People thus turns to neural radiance fields (NeRFs) \cite{mildenhall2020nerf}, as it offers a more promising approach to modeling large scenes.
With careful design~\cite{chen2022tensorf,fridovich2022plenoxels,muller2022instant,sun2022direct,lin2022neurmips}, NeRF is able to render free-viewpoint, photo-realistic images efficiently. 
However, crafting an interactive environment entails more than just creating a visually high-fidelity digital twin; it also involves building a {\bf physically plausible}, {\bf immersive}, {\bf real-time} and importantly, {\bf interactive} world tailored to user experiences. 
Furthermore, we expect such a virtual world to be {\bf compatible with real-time interaction interfaces} such as common game engines.
Despite its promise, the use of NeRF to create interactive environments from real-world videos remains largely unexplored.

In this paper, we introduce Video2Game, a novel approach to automatically converting a video of a scene into a {realistic and interactive virtual} environment. 
Given a video as input, we first construct a NeRF that can effectively capture the geometric and visual information of a (large-scale, unbounded) scene. 
Then we distill the NeRF into a game engine-compatible, neural textured mesh.  This significantly improves the rendering efficiency while maintains the overall quality.
To model the interactions among the objects, we further decompose the scene into individual actionable entities and equip them with respective physics model.
Finally, we import our automatically generated assets into a WebGL-based game engine and create a playable game. 
The resulting virtual environment is photo-realistic, interactive, and runs in {real-time. See Fig.~\ref{fig:teaser} for demonstration.}  
In summary, our key contributions are:
\vspace{-4px}
\begin{itemize}
    \item A novel 3D modeling algorithm for real-time, free-viewpoint rendering and physical simulation, surpassing state-of-the-art NeRF baking methods with added rigid-body physics for enhanced simulation.
    \item An automated game-creation framework to transform a scene video into an interactive, realistic environment, compatible with current game engines.
\end{itemize}

\section{Related Works}
Given a single video, we aim to create a real-time, interactive game where the agents (\eg, the character, the car) can navigate and explore the reconstructed digital world, interact with objects in the scene (\eg, collision and manipulate objects), and achieve their respective tasks (\eg, collecting coins, shooting targets).
We draw inspirations from multiple areas and combine the best of all. In this section, we will briefly review those closely related areas which forms the foundation of our work.

\noindent \textbf{Novel view synthesis (NVS): }
Our work builds upon the success of novel view synthesis \cite{chen1993view,levoy1996light,szeliski1998stereo,heigl1999plenoptic},
which is crucial for our game since it enables the agents to move freely and view the reconstructed world seamlessly from various perspectives.
Among all these approaches \cite{zhou2018stereo,tulsiani2018layer,srinivasan2019pushing,hu2021worldsheet,zuo2022view}, we exploit neural radiance field (NeRF) \cite{mildenhall2020nerf} as our underlying representation.
NeRF has emerged as one of the most promising tools in NVS since its introduction \cite{niemeyer2021giraffe,park2021nerfies,park2021hypernerf}, and has 
great performance across a wide range of scenarios \cite{zhang2020nerf++,li2022climatenerf,rematas2022urban,yang2023unisim}. 
For instance, it can be easily extended to handle various challenging real-world scenarios such as learning from noisy camera poses \cite{lin2021barf,wang2021nerf}, reflectance modeling for photo-realistic relighting \cite{zhang2021nerfactor,verbin2022ref}, and real-time rendering \cite{lin2022neurmips,chen2022mobilenerf,yariv2023bakedsdf, reiser2023merf, tang2022nerf2mesh}.
In this work, we combine recent advances in NeRF with {\it physics modeling} to build an immersive digital world from one single video, moving from \emph{passive} NVS to our complete solution for \emph{embodied} world modeling where agents can \emph{actively} explore and interact with the scene.

\noindent \textbf{Controllable video generation: }
Using different control signals to manipulate the output of a visual model has garnered great interest in the community. 
This has had a profound impact on content creation \cite{singer2022make,singer2023text}, digital editing \cite{bar2022text2live,lee2023shape}, and simulation \cite{kim2020learning,kim2021drivegan, lin2023urbanir}. 
One could also leverage large foundation models to control video content using text \cite{singer2022make,singer2023text}. 
However, they lack fine-grained and real-time control over the generated content.
Alternatively, training (conditional) generative models for each scene enables better disentanglement of dynamics (\eg, foreground \emph{vs.} background) and supports better control. 
For instance, one can represent a self-driving scene \cite{kim2021drivegan} or a Pacman game \cite{kim2020learning} as latent codes and generate video frames based on control inputs with a neural network.
One can also learn to control the players within tennis games \cite{zhang2021vid2player,menapace2021playable,menapace2022playable, zhang2023vid2player3d}.
Our work falls under the second line of research, where the model takes user control signals (\eg, a keystroke from the keyboard) as input and responds by rendering a new scene.
However, instead of focusing on a specific scene (\eg, tennis games), we have developed a pipeline that allows the creation of a playable environment from a single video of a generic scene.
Additionally, we model everything in 3D, which enables us to effectively capture not only view-dependent appearance but also physical interactions among rigid-body equipped objects. 
Importantly, we adopt a neural representation compatible with graphics engines, enabling users to play the entire game in their browser at an interactive rate.

\noindent \textbf{Data-driven simulation: }
Building a realistic simulation environment has been a longstanding challenge.
\cite{khosla1985parameter,wymann2000torcs,todorov2012mujoco,dosovitskiy2017carla}.
While it's promising, 
we come close to mirror the real world only in recent years
\cite{yang2020surfelgan,manivasagam2020lidarsim,chen2021geosim,amini2022vista,son2022differentiable,yang2023unisim,lu2023urban}.
The key insight of these work is to build models by leveraging real-world data.
Our work closely aligns with this line of research on building high-fidelity simulators from real-world data, with a few key differences.
First, existing works mainly focus on \emph{offline} training and evaluation \cite{yang2020surfelgan,manivasagam2020lidarsim,amini2022vista,yang2023unisim}, whereas our system runs at an interactive rate and allows for \emph{online}, \emph{real-time} control. 
Second, some existing works\cite{manivasagam2020lidarsim,zyrianov2022learning,ultralidar,liu2023real} need additional data modality like LiDAR point clouds for geometry reconstruction, but RGB video is all we need. 
Third, most photo-realistic simulators don't model physical interactions. However, we supports various physics modeling and allows agents to interact with the environment.
Last, existing simulators are typically resource-intensive
, while our system is lightweight and can be easily accessible in common engines.

\begin{table}[t]
    \centering
    \setlength\tabcolsep{0.05em} %
    \resizebox{\linewidth}{!}{
    \begin{tabular}{c}

      \includegraphics[width=\linewidth]{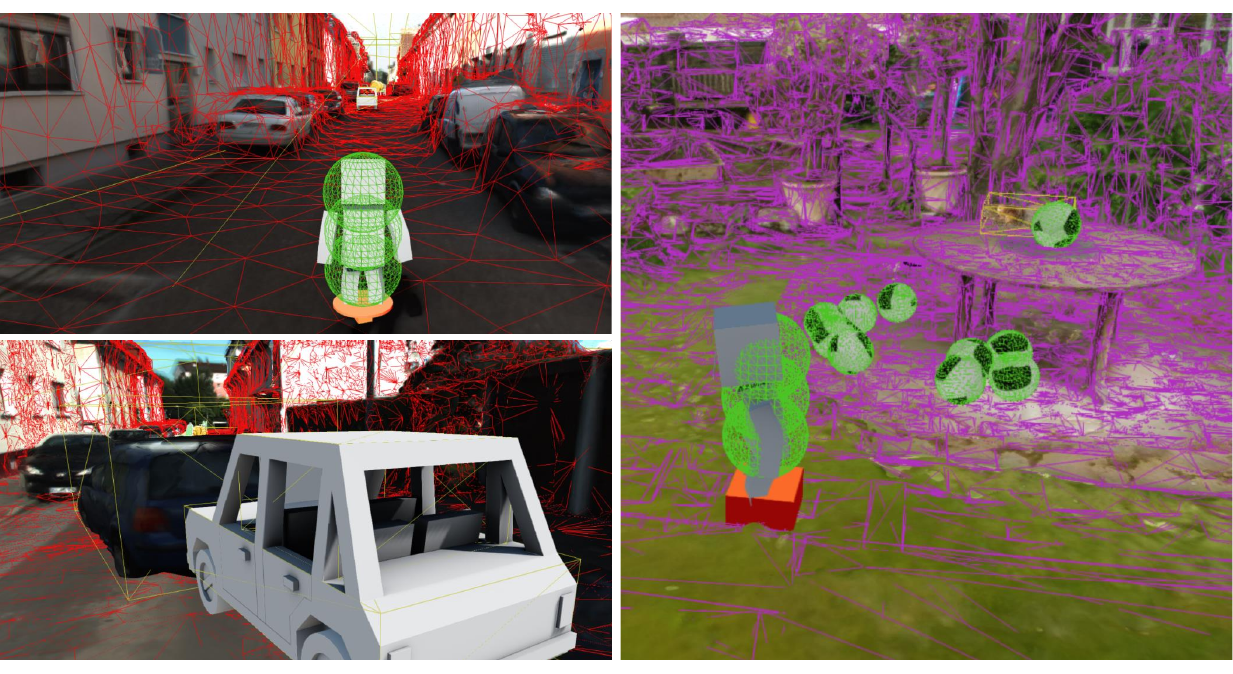}
        
    \end{tabular}
    }
    \captionof{figure}{\textbf{Visualization of automatically computed collision geometry:}  Sphere collider (green), box collider (yellow), convex polygon collider (purple) and trimesh collider (red). 
    }
    \label{fig:collision}
\end{table}

\section{Video2Game}
Given a sequence of images or a video of a scene, our goal is to construct an \emph{interactable} and \emph{actionable} digital twin, 
upon which we can build real-time, interactive games or realistic (sensor) simulators.
Based on the observations that prevalent approaches to constructing digital replica mainly focus on visual appearance and ignore the underlying physical interactions, we carefully design our system such that it can not only produce high-quality rendering across viewpoints, but also support the modeling of physical actions (\eg, navigation, collision, manipulation, etc).
At the heart of our systems is a compositional implicit-explicit 3D representation that is effective and efficient for both sensor and physics simulation.
By decomposing the world into individual entities, we can better model and manipulate their physical properties (\eg, specularity, mass, friction), and simulate the outcomes of interactions more effectively.

We start by introducing a NeRF model that can effectively capture the geometric and visual information of a large-scale, unbounded scene (Sec.~\ref{sec:nerf}). %
Next, we present an approach to convert the NeRF into a game-engine compatible mesh with neural texture maps, significantly improving the rendering efficiency while maintaining the quality (Sec.~\ref{sec:baking}). 
To enable physical interactions, we further decompose the scene into individual actionable entities and equip them with respective physics models (Sec.~\ref{sec:physics}).
Finally, we describe how we integrate our interactive environment into a WebGL-based game engine, allowing users to play and interact with the virtual world in real time on their personal browser.
Fig.~\ref{fig:overview} provides an overview of our proposed framework.

\begin{table*}[t]
    \centering
    \setlength\tabcolsep{0.05em} %
    \resizebox{\textwidth}{!}{
    \begin{tabular}{lcccc}

        \raisebox{5mm}[0pt][0pt]{\rotatebox[origin=c]{90}{RGB}} & \includegraphics[width=0.24\textwidth]{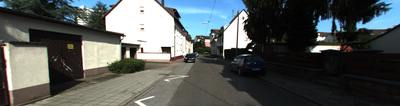} & \includegraphics[width=0.24\textwidth]{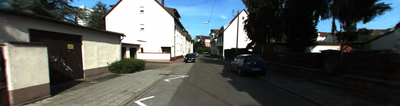} & \includegraphics[width=0.24\textwidth]{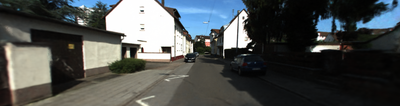} & \includegraphics[width=0.24\textwidth]{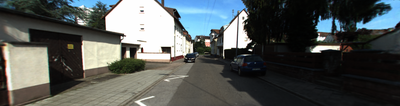} \\

        \raisebox{5mm}[0pt][0pt]{\rotatebox[origin=c]{90}{Depth}} & 
        \includegraphics[width=0.24\textwidth]{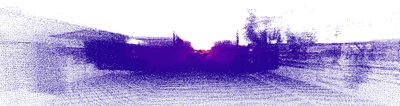} 
        & \includegraphics[width=0.24\textwidth]{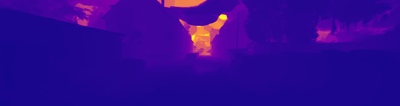} & \includegraphics[width=0.24\textwidth]{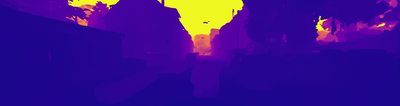} & \includegraphics[width=0.24\textwidth]{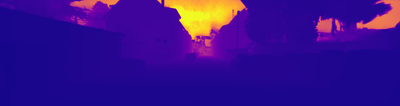} \\

        \raisebox{5mm}[0pt][0pt]{\rotatebox[origin=c]{90}{Normal}} & 
        \includegraphics[width=0.24\textwidth]{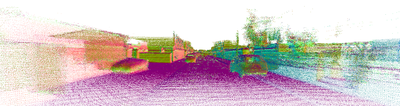} 
        & \includegraphics[width=0.24\textwidth]{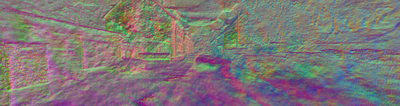} & \includegraphics[width=0.24\textwidth]{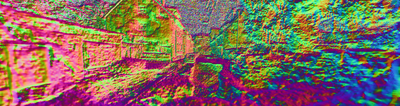} & \includegraphics[width=0.24\textwidth]{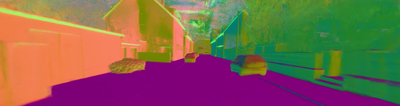} \\
       
        & Ground Truth & Instant-NGP & Nerfacto & Ours
    \end{tabular}
    }
    \captionof{figure}{{\bf Qualitative comparisons among NeRF models.} The rendering quality of our base NeRF is superior to baselines, and with leveraging monocular cues, we substantially improve rendered geometry compared to other baselines. This significantly facilitates NeRF baking in subsequent stages. Here we consider depths measured by LiDAR point cloud in KITTI-360 and compute normals based on it.
    }
    \label{tab:nerf_rendering_qualti}
\end{table*}

\subsection{Large-scale NeRF}
\label{sec:nerf}

\paragraph{Preliminaries:} 
Instant-NGP~\cite{muller2022instant} is a notable variant of NeRF,
which represents the radiance field with a combination of spatial hash-based voxels and neural networks:
$\bc, \sigma = F_\theta(\mathbf{x}, \bd; \Phi) = \texttt{MLP}_\theta(\texttt{It}(\bx, \Phi), \bd)$.
Given a 3D point $\bx \in \bbR^3$ and a camera direction $\bd \in \bbR^2$ as input, Instant-NGP first interpolate the point feature $\texttt{It}(\bx, \Phi)$ from the adjacent voxel features $\Phi$. Then the point feature and the camera direction are fed into a light-weight multi-layer perception (MLP) to predict the color $\bc \in \bbR^3$ and density $\sigma \in \bbR^{+}$. 
\zhihao{To render the scene appearance,}
we first cast a ray $\br(t) = \bo + t\bd$ from the camera center $\bo$ through the pixel center in direction $\bd$, and sample a set of 3D points $\{\bx_i\}$ along the ray.
We then query their respective color $\{\bc_i\}$ and density $\{\sigma_i\}$ and obtain the color of the pixel through alpha-composition: $\bC_\text{NeRF}(\br) = \sum_i w_i \bc_i$. 
Similarly, the expected depth can be computed by: $\bD_\text{NeRF}(\br) = \sum_i w_i t_i$. Here, $w_i$ indicates the blending weight that is derived from the densities $\{\sigma_i\}$. We refer the readers to \cite{mildenhall2020nerf} for more details.
To learn the voxel features $\Phi$ and the MLP weights $\theta$, we compute the difference between the ground truth color and the rendered color:
$\mathcal{L}_\text{rgb} = \sum_{\br} \lVert\bC_\text{GT}(\br) - \bC_\text{NeRF}(\br) \rVert_2^2$.

\vspace{-15px}
\paragraph{Large-scale NeRF: }
While Instant-NGP \cite{muller2022instant} has shown promising results on densely observed and bounded scenes, its performance starts to degrade when extending to \hongchi{sparsely-captured}, large-scale, unbounded environments.
To mitigate these issues, we propose several enhancements:
\vspace{-10px}
\begin{equation}
\bc, \sigma, s, \bn = F_\theta(\mathbf{x}, \bd; \Phi) 
= \texttt{MLP}_\theta(\texttt{It}(\texttt{Ct}(\bx), \Phi), \bd).
\label{eq:our-nerf}
\end{equation}
\hongchi{First of all, we exploit the contraction function $\texttt{Ct}(\bx)$ ~\cite{barron2022mip} to map the unbounded coordinates into a bounded region.
In addition to radiance and density, we predict the semantics $s$ and the surface normal $\bn$ of the 3D points, guided with 2D priors to better regularize the scene geometry.
Furthermore, we divide large-scale scenes into several blocks~\cite{tancik2022block} to capture the fine-grained details. We now describe these enhancements in more details.}

\vspace{-15px}
\paragraph{Depth: }
High-quality geometry is critical for modeling physical interactions. 
Inspired by MonoSDF \cite{yu2022monosdf}, we leverage off-the-shelf monocular depth estimators \cite{eftekhar2021omnidata, kar20223d} to guide and improve the underlying geometry. 
We first predict the depth of the scene from rendered RGB images. Then we minimize the discrepancy between the rendered depth and the predicted depth via $ \mathcal{L}_\text{depth} =\sum_\br \| (a\bD_{\text{NeRF}}(\br) + b) - \bD_{\text{mono}}(\br) \|^{2}_2 $, where $a$ and $b$ are the scale and shift that aligns the two distribution \cite{ranftl2020towards}.

\vspace{-10px}
\paragraph{Surface normals: } 
Similar to depth, we encourage the normal estimated from NeRF to be consistent with the normal predicted by the off-the-shelf estimator \shortcite{eftekhar2021omnidata, kar20223d}. 
The normal of a 3D point $\bx_i$ can be either analytically derived from the estimated density $\bn_i = (1 - \frac{\nabla_\bx \sigma_i} {\| \nabla \sigma_i \| })$, or predicted by the MLP header as in Eq. \ref{eq:our-nerf}. We could aggregate them via volume render: $\mathbf{N}(\br) = \sum_i \bw_i \bn_i$.
Empirically we find that adopting both normals and promoting their mutual consistency works the best, since the MLP header offers more flexibility.
We thus employ 
$\mathcal{L}_\text{normal} = \| \mathbf{N}_\text{mlp}(\br) - \mathbf{N}_\text{mono}(\br)\|_2^2 + \| \mathbf{N}_\text{mlp}(\br) - \mathbf{N}_\text{density}(\br)\|_2^2$.

\vspace{-10px}
\paragraph{Semantics: }
We also predict semantic logits for each sampled 3D points with our MLP. This helps us capture the correlation between semantics and geometry~\cite{zhi2021inplace, li2022climatenerf}. We render the semantic map with volume rendering $\bS_{\text{NeRF}}(\br) = \sum_i \bw_i \bs_i$ and compute the cross-entropy with that of a 2D segmentation model
$\mathcal{L}_{\text{semantics}} = \texttt{CE}\left(\bS_{\text{mono}}, \bS_{\text{NeRF}}\right).$

\vspace{-10px}
\paragraph{Regularization: } 
We additionally adopt two regularization terms. To reduce floaters in the scene, for each randomly sampled 3D point $\bx$, we penalize its density by $\mathcal{L}_\text{sp} = \sum 1 - \text{exp}(-\alpha\sigma(\bx))$, where $\alpha > 0$~\cite{yu2021plenoctrees}.
For each sky pixel (which we derived from the semantic MLP), we encourage its depth $\bD_{\text{NeRF}}(\br^{\text{sky}})$ to be as far as possible. The loss is defined as:
$\mathcal{L}_\text{sky} = \sum_{\br^{\text{sky}}} \text{exp}(-\bD_{\text{NeRF}}(\br^{\text{sky}}))$.

\vspace{-10px}
\paragraph{Blocking:} 
\label{sec:block}
Capitalizing on a single Instant-NGP to cover an extraordinarily large scene such as KITTI-360~\cite{liao2022kitti} would often lead to inferior results. We thus adopt a strategy akin to BlockNeRF~\cite{tancik2022block} where we divided the whole scene into numerous blocks and model each region with a separate Instant-NGP. Adjacent regions maintain substantial overlaps to ensure smooth transition.

\vspace{-10px}
\paragraph{Learning:}
We jointly optimize the voxel feature $\Phi$  and the MLP weights $\theta$ by minimizing the following loss: 
\begin{equation}
    \mathcal{L}^{\text{NeRF}}_\text{total} = \mathcal{L}_\text{rgb} + \mathcal{L}_\text{normal} + \mathcal{L}_\text{semantics} + \mathcal{L}_\text{depth} + \mathcal{L}_\text{sky} + \mathcal{L}_\text{sp}
\end{equation}

\begin{table*}[t]

\centering
\resizebox{\textwidth}{!}{
\begin{tabular}{lcccccccccc}
\toprule
{\multirow{2}{*}{Method}}  & {\multirow{2}{*}{Representation}} & \multicolumn{3}{c}{KITTI-360} & \multicolumn{3}{c}{Gardenvase}  & \multicolumn{3}{c}{Interactive Compatibility}                                \\ 
 & &PSNR$\uparrow$    & SSIM$\uparrow$   & LPIPS$\downarrow$  & PSNR$\uparrow$      & SSIM$\uparrow$     & LPIPS$\downarrow$   & Real time &  Rigid-body physics & Scene decomposition       \\ \midrule
InstantNGP~\cite{muller2022instant} & \multirow{3}{*}{Volume} & 27.46  &  0.853 & 0.165  &   25.90   &   0.757  &    0.191   &  \xmark  &  \xmark  &  \xmark  \\
Nerfacto~\cite{nerfstudio}   & & 23.20 & 0.763 &  0.238 & 22.16  &  0.517   &  0.283    &  \xmark  &  \xmark  &  \xmark            \\
Video2Game &  & \textbf{27.62} &  \textbf{0.871}  &  \textbf{0.131} &  \textbf{26.57}    & \textbf{0.815} & \textbf{0.143}  &  \xmark  &  \xmark  &  \xmark        \\ \midrule
Gauss. Spl.~\cite{kerbl20233d} &  Points & 17.85  & 0.615 &  0.428 & {27.50}  & {0.858}  &     {0.099}    & \cmark    &  \xmark  &  \xmark   \\ \midrule
MobileNeRF~\cite{chen2022mobilenerf} & \multirow{3}{*}{Mesh} & 19.67 & 0.627 & 0.452  & 22.80  & 0.505  & 0.365 & \cmark &  \xmark  &  \xmark   \\
BakedSDF*~\cite{yariv2023bakedsdf} &  & 22.37 & 0.757  & 0.302 & 22.68  & \textbf{0.514}  & 0.369   & \cmark  &    \cmark      &  \xmark       \\
Video2Game  &    & \textbf{23.35}  & \textbf{0.765} & \textbf{0.246}  & \textbf{22.81}  & 0.508  &   \textbf{0.363}  & \cmark   &  \cmark  &  \cmark   \\ \bottomrule
\end{tabular}
}
\caption{\textbf{Quantitative results on novel view synthesis and interactive compatibility analysis.} {Video2Game produces better or comparable results across scenes, suggesting the effectiveness of our NeRF and mesh model.
The performance improves the most when tackling unbounded, large-scale scenes in KITTI-360. 
We note that existing NeRFs cannot reach the interactive rate required for real-time games. 
While point-based rendering significantly improves the speed, it does not support rigid body physics. BakedSDF \cite{yariv2023bakedsdf} represents the whole scene with one single mesh, thus does not support object-level interactions.}} %
\label{tab:nerf_rendering_quanti}

\end{table*}

\subsection{NeRF Baking}
\label{sec:baking}
We aim to create a digital replica that users (or agents) can freely explore and act upon in real time.
Although our large-scale NeRF effectively renders high-quality images and geometry, its efficiency is limited by the computational costs associated with sampling 3D points.
The underlying volume density representation further complicates the problem. 
For instance, it's unclear how to define physical interaction with such a representation (\eg, defining collision).
The representation is also not compatible with common graphics engines.
While recent software, such as the NeRFStudio Blender plugin and LumaAI Unreal add-on, has made some strides, their interaction capabilities and scene geometry quality are still not optimal for real-time user engagement, especially when the scene is large and the observations are relatively sparse.
To overcome these challenges, we draw inspiration from recent NeRF meshing advancements and present a novel NeRF baking framework that  efficiently transforms our NeRF representation into a game-engine compatible mesh.
As we will show in Sec. \ref{sec:exp}, 
this conversion greatly enhances rendering efficiency while preserving quality and facilitates physical interactions.

\vspace{-8px}
\paragraph{Mesh representation: }
Our mesh $\mathcal{M} = (\bV, \bF, \bT)$ comprises vertices $\bV \in \bbR^{|V|\times3}$, faces $\bF \in \mathbb{N}^{|F|\times3}$ and a UV neural texture map $\bT \in \bbR^{H\times W\times6}$. Following \cite{tang2022nerf2mesh}, we store the base color in the first three dimension of $\bT$, and encode the specular feature in the rest. 
The initial mesh topology are obtained by marching cubes in the NeRF density field. 
We further prune the invisible faces. conduct mesh decimation and edge length regularization.
The UV coordinate of each vertex is calculated via xatlas~\cite{xatlas}.

\vspace{-8px}

\paragraph{Rendering:} 
We leverage differentiable renderers \cite{Laine2020diffrast} to render our mesh into RGB images $\bC_\text{R}$ and depth maps $\bD_\text{R}$. Specifically, we first rasterize the mesh into screen space and obtain the UV coordinate for each pixel $i$. Then we sample the corresponding texture feature $\bT_i = [\bB_i; \bS_i]$ and feed it into our customized shader. Finally, the shader computes the sum of the view-independent base color $\mathbf{B}_i$ and the view-dependent MLP $\texttt{MLP}_\theta^\text{shader}(\mathbf{S}_i, \bd_i)$:
\begin{equation}
\label{eqn:tr}
\bC_\text{R} = \mathbf{B}_i + \texttt{MLP}_\theta^\text{shader}(\mathbf{S}_i, \bd_i).
\end{equation}
The MLP is lightweight and can be baked in GLSL.

\vspace{-8px}
\paragraph{Learning: }
We train the shader MLP $\texttt{MLP}_\theta^\text{shader}$ and the neural texture map $\bT$ by minimizing the color difference between the mesh and the ground truth, and the geometry difference between the mesh and the NeRF model:
\begin{equation}
\small
\mathcal{L}^\text{mesh}_{\bT, \theta} = \sum_{\br}  \| \bC_\text{R}(\br) - \bC_\text{GT}(\br) \| + \| \bD_\text{R}(\br) - \bD_\text{NeRF}(\br) \|.
\end{equation}

\vspace{-13px}

\paragraph{Anti-aliasing:} 
Common differentiable rasterizers only take the center of each pixel into account.
This may lead to aliasing in the learned texture map.
To resolve this issue, we randomly perturb the optical center of the camera by 0.5 pixels along each axis at every training step.
This ensure all the regions within a pixel get rasterized.

\vspace{-8px}

\paragraph{Relationship to existing work:} 
Our approach is closely related to recent work on NeRF meshing \cite{chen2022mobilenerf, reiser2023merf, tang2022nerf2mesh, yariv2023bakedsdf}, but there exist key differences. While MobileNeRF~\cite{chen2022mobilenerf} also adopts an explicit mesh with neural textures, they mainly capitalize on planar primitives. The quality of the reconstructed mesh is thus inferior. BakedSDF~\cite{yariv2023bakedsdf} offers excellent runtime and rendering quality, but their vertex coloring approach has limited resolution for large scenes. NeRF2Mesh~\cite{tang2022nerf2mesh} lacks depth distillation and doesn't adopt contraction space for unbounded scenes. 
They also have a sophisticated multi-stage training and multi-resolution refinement process. 
Finally, MeRF~\cite{reiser2023merf}, though efficient, still relies on volume-rendering.

\subsection{Representation for Physical Interaction}
\label{sec:physics}
Our mesh model facilitates efficient novel-view rendering in real time and allows for basic \emph{rigid-body} physical interactions. For example, the explicit mesh structure permits an agent to ``stand'' on the ground.
Nevertheless, beyond navigation, an agent should be capable of performing various actions including collision and manipulation.
Furthermore, a scene comprises not only the background but also interactable foreground objects, each possessing unique physical properties. 
For instance, a street-bound car is much heavier than a flower vase. When struck by another object, a car may barely move but the vase may fall and shatter.
To enhance physical interaction realism, we decompose the scene into discrete, actionable entities, each endowed with specific physical characteristics (\eg, mass, friction). This approach, in conjunction with \emph{rigid-body} physics, allows for the effective simulation 
that adheres to physical laws.

\vspace{-10px}

\paragraph{Scene decomposition: }
Directly editing and decomposing a mesh is extremely difficult due to topology change.
Fortunately, neural fields are inherently compositional in 3D. 
By identifying the objects each spatial region belongs to, we can use neural fields to guide the decomposition of the mesh.
Specifically, we sample a 3D point $\bx_i$ within each voxel $i$ and determine its semantic category either through the predicted semantic logits $s_i$ or by verifying whether the point is within a specified bounding box. 
The process is repeated for all voxels to segment the entire scene.
Then, for each object, we perform NeRF meshing individually, setting the density of the remaining areas to zero.
The intersections between objects are automatically resolved by  marching cube.
Finally, we initialize the neural texture of these new, individual meshes from the original mesh model. For newly created faces, we employ nearest neighbor inpainting on the neural texture map, which empirically yields satisfactory results. 
Fig. \ref{fig:teaser} shows an example where a vase is separated from a table. The middle of the table is original occluded yet we are able to  maintain high-quality rendering.

\vspace{-10px}

\paragraph{Physical parameters reasoning: }
The next step is to equip decomposed individual meshes with various physics-related attributes so that we can effectively model and simulate their physical dynamics. 
In this work, we focus on rigid body physics, where each entity $i$ is represented by a collision geometry $col_i$, mass $m_i$, and friction parameters $f_i$.
We support fours types of collision geometry with different levels of complexity and efficiency: box, sphere, convex polygon, and triangle mesh. 
Depending on the object and the task of interest, one can select the most suitable collision check for them.
For other physical parameters (\eg mass, friction), one can either set them manually or query large language models (LLMs) for an estimation.

\begin{table*}[t]
    \centering
    \setlength\tabcolsep{0.05em} %
    \resizebox{0.9\textwidth}{!}{
    \begin{tabular}{cccc}
        \includegraphics[width=0.24\textwidth]{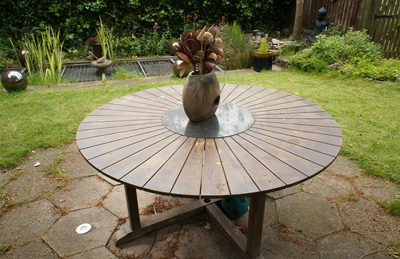} & 
        \includegraphics[width=0.24\textwidth]{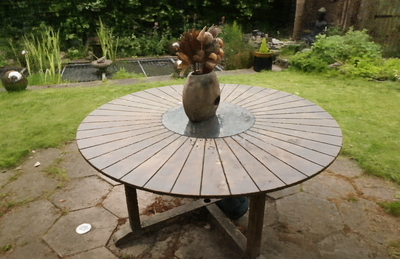} & 
        \includegraphics[width=0.24\textwidth]{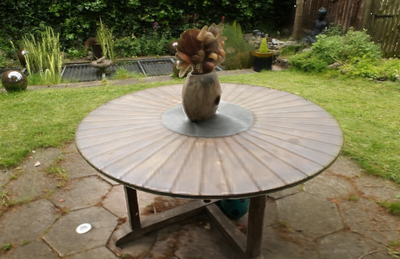} & 
        \includegraphics[width=0.24\textwidth]{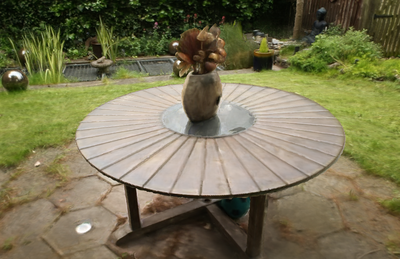} \\
        
        \includegraphics[width=0.24\textwidth]{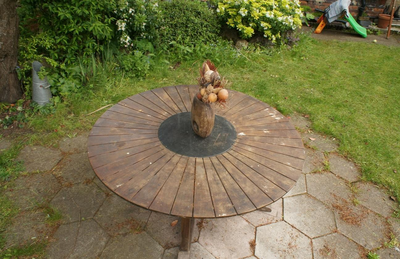} & 
        \includegraphics[width=0.24\textwidth]{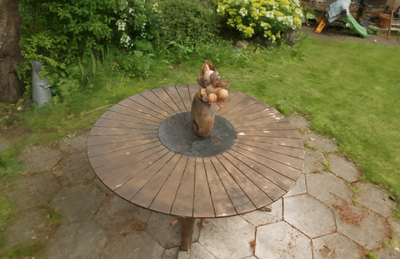} & 
        \includegraphics[width=0.24\textwidth]{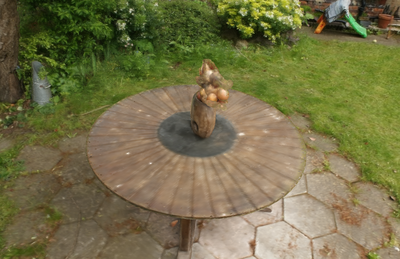} & 
        \includegraphics[width=0.24\textwidth]{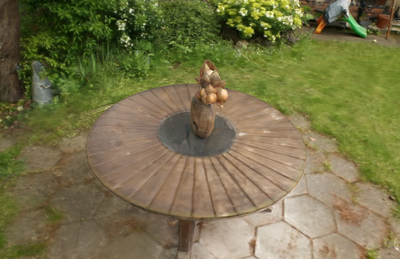} \\
         G.T. & MobileNeRF~\cite{chen2022mobilenerf} & BakedSDF~\cite{yariv2023bakedsdf} & Ours
       
    \end{tabular}
    }
    \captionof{figure}{{\bf Qualitative comparisons among mesh models.} We compare our mesh rendering method with others in Garden scene~\cite{barron2022mip}. 
    }
    \label{tab:main_mesh_qualitative}
\end{table*}

\vspace{-10px}

\paragraph{Physical interactions: }
Rigid body dynamics, while simple, can support a variety of interactions. 
With the collision check, an user/agent can easily navigate through the environment while respecting the geometry of the scene.
The agents will no longer be stuck in a road or cut through a wall.
It also allows the agent to interact with the objects in the scene.
For instance, one can push the objects towards the location of interest. The object movement will be determined by its mass and other physical properties such as the friction.
We can also manipulate the objects by adopting a magnet grasper, following AI2-Thor \cite{kolve2022ai2thor}.
This opens the avenue towards automatic creation of realistic, interactive virtual environment for robot learning.

\begin{table}[t]
    \centering
    \setlength\tabcolsep{0.05em} %
    \resizebox{0.5\textwidth}{!}{
    \begin{tabular}{cc}
          \includegraphics[width=0.265\textwidth]{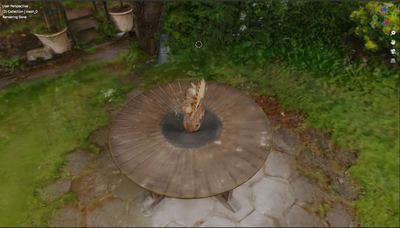} & 
          \includegraphics[width=0.25\textwidth]{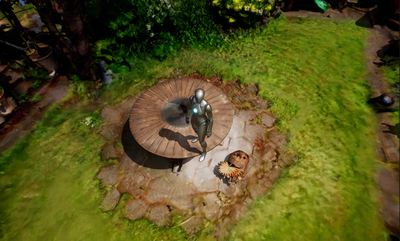} \\
          Blender & Unreal   \\ 
    \end{tabular}
    }
    \vspace{-2mm}
    \captionof{figure}{{\bf Video2Game in Blender and Unreal Engine.}}
    \label{tab:platform}
\end{table}

\subsection{Interactive Environment}
\label{sec:game}
We deploy our interactive environment within a real-time, browser-based game engine. 
We manage the underlying logic and assets using Sketchbook~\cite{Sketchbook}, a Game Engine based on Three.js that leverages WebGL \cite{WebGL} for rendering. 
This combination ensures high efficiency while offering the flexibility and sophistication required for intricate rendering tasks. 
It also allows us to easily integrate content from different scenes together.
We have further extended Sketchbook's capabilities by implementing a GLSL-based shader \cite{GLSL}. 
This enables real-time computation of our MLP-based specular shader during deployment. 
For physics simulation, we use Cannon.js~\cite{Cannon}, which assures realism and efficiency in the motion within our interactive environment.
It supports not only rigid body dynamics but also more sophisticated modeling techniques. 
For example, we can pre-compute the fracturing effect for dynamic objects. Upon experiencing a significant force, these objects are realistically simulated by the real-time physics engine, which handles the interactions between the fractured pieces and the rest of the scene, such as their falling and settling on the ground. 
\hongchi{Besides browser-based engine, the virtual environments from Video2Game pipeline could be also integrated into both \textbf{Blender}~\cite{blender} and \textbf{Unreal} engines~\cite{unrealengine} (see Fig.~\ref{tab:platform}).}

\section{Experiments}
\label{sec:exp}
We begin by presenting our experimental setup, followed by a comparison of our model with state-of-the-art approaches. Next, we conduct an extensive analysis of our model's distinctive features and design choices. Then we demonstrate how we constructed a web browser-compatible game capable of delivering a smooth interactive experience exceeding 100 frames per second (FPS), all derived from a single video source. Finally, we showcase the capabilities of our model in robot simulation through two demonstrations.

\begin{table}[t]
    \centering
    \resizebox{0.38\textwidth}{!}{
    \begin{tabular}{lccc}
    \toprule
       Method &  Outlier-\%$\downarrow$ & RMSE$\downarrow$  & MAE$\downarrow$ \\
        \midrule
         Instant-NGP~\cite{muller2022instant} & 22.89 & 4.300 & 1.577 \\
         Nerfacto~\cite{nerfstudio} & 50.95 & 8.007 & 2.863 \\
         Gauss. Spl.~\cite{kerbl20233d} & 91.08 & 11.768 & 8.797 \\
         BakedSDF* (offline)~\cite{yariv2023bakedsdf} & 43.78 & 5.936 & 2.509 \\
        \name  \hspace{1mm}(Our NeRF) &  \textbf{13.23} & \textbf{3.028} & \textbf{1.041} \\ %
\bottomrule
    \end{tabular}}
    \caption{\textbf{Quantitative evaluation on NeRF geometry.} {Our NeRF renders significantly more accurate depth compared with the baselines. The unit is meter and the outlier threshold is 1.5 meters.}}
    \label{tab:nerf_rendering_quanti_geo}
    
\end{table}

\begin{table}[t]
    \vspace{-2mm}
    \centering
    
    \resizebox{0.48\textwidth}{!}{
    \begin{tabular}{p{0.2\textwidth}cccccc}
    \toprule
        \multirow{2}{*}{Method} & \multicolumn{3}{c|}{Volume Rendering} & \multicolumn{3}{c}{Mesh Rastization}  \\
        & PSNR$\uparrow$ & SSIM$\uparrow$ & LPIPS$\downarrow$  & PSNR$\uparrow$ & SSIM$\uparrow$ & LPIPS$\downarrow$ \\
        \midrule
         Vanilla NGP  & 27.46 & 0.853 & 0.165 & 22.54 & 0.716 & 0.350   \\
         + Regularization terms  & 27.52 & 0.861  & 0.157   & 22.97 & 0.732 & 0.303 \\
         + Monocular cues  & \textbf{27.62} & \textbf{0.871} & \textbf{0.131}  & \textbf{23.35} & \textbf{0.765} & \textbf{0.246}  \\\bottomrule
         
    \end{tabular}
    }

    \caption{\textbf{Ablation studies}{ on KITTI-360.
    }}
    \label{tab:abalation_table}
    
\end{table}

\vspace{-5px}
\subsection{Setup}
\noindent \textbf{Dataset: }
\label{sec:data}
We evaluate the effectiveness of Video2Game across three distinct scenes in various scenarios, including ``Gardenvase"~\cite{barron2022mip}, an outdoor object-centric scene; the KITTI-360 dataset~\cite{liao2022kitti}, a large-scale self-driving scene with a sequence that forms a closed loop, suitable for car-racing and Temple Run-like games; and finally, an indoor scene from the VR-NeRF~\cite{VRNeRF} dataset, showcasing the potential for robot simulations.

\noindent \textbf{Metrics: }
To evaluate the quality of the rendered images, we adopt PSNR, SSIM, and LPIPS \cite{zhang2018unreasonable}.
For geometry reconstruction, we evaluate with LiDAR point cloud in KITTI-360 dataset. Root mean square deviation (RMSE), mean absolute error (MAE), and outlier rate are applied to measure the disparity existing in estimated geometry. 

\noindent \textbf{Our model: } {For NeRF, we adopt hashgrid encoding \cite{tiny-cuda-nn} and two-layer MLP for each header. For textured mesh, we conduct marching cubes on the NeRF and post-process it to a fixed precision. We set the texture image size to 4096x4096. For GLSL shader, we design a light-weight two-layer MLP, which enables efficient real-time rendering. For KITTI-360 (see Sec.~\ref{sec:block}), we divide the whole scene into 16 blocks and create a skydome mesh for the sky. 
}

\noindent \textbf{Baselines: }
To evaluate the visual and geometry quality of our model, we compare against SOTA approaches in neural rendering and neural reconstruction. 
\textbf{Instant-NGP} \cite{muller2022instant} is a NeRF-based method that exploits multi-resolution hashing encoding. 
\textbf{Nerfacto} \cite{nerfstudio} extends the classic NeRF with learnable volumetric sampling and appearance embedding.
\textbf{3D Gaussian Splatting} \cite{kerbl20233d} leverages 3D Gaussians and achieves fast training and rendering.
\textbf{MobileNeRF} \cite{chen2022mobilenerf} adopts a hybrid NeRF-mesh representation. It can be baked into a texture map and enable real-time rendering.
\textbf{BakedSDF} \cite{yariv2023bakedsdf} adopts a volume-surface scene representation. It models view-dependent appearance efficiently by baking spherical Gaussians into the mesh.

\subsection{Experimental results}
\label{sect:exp_vol_render}
\noindent \textbf{Novel view synthesis: }
{Tab. \ref{tab:nerf_rendering_quanti} shows the rendering performance and interactive compatibility of our model against the baselines on KITTI-360  \cite{liao2022kitti} and Gardenvase \cite{barron2022mip}. Our NeRF achieves superior performance when compared to state-of-the-art neural volume render approaches across different scenes. Though \cite{kerbl20233d}  performs best in Gardenvase, it fails to handle the sparse camera settings in KITTI-360, where it learns bad 3D orientations of Gaussians. Our baked mesh outperforms other mesh rendering baselines significantly in KITTI-360 and performs similarly in Gardenvase \hongchi{as shown in  Fig.~\ref{tab:main_mesh_qualitative}}. Additionally, our pipeline has the highest interactive compatibility among all baselines.}

\noindent \textbf{Geometry reconstruction: }
Our model performs significantly better than the baseline regarding geometry accuracy (see Tab.~\ref{tab:nerf_rendering_quanti_geo}). We provide some qualitative results in Fig. \ref{tab:nerf_rendering_qualti}, demonstrating that our model can generate high-quality depth maps and surface normals, whereas those produced by the baselines contain more noise.

\noindent \textbf{Ablation study: }
To understand the contribution of each component in our model, we begin with the basic Instant-NGP~\cite{muller2022instant} and sequentially reintroduce other components. The results in Tab.~\ref{tab:abalation_table} show that our regularization and monocular cues improve the quality of both volume rendering in NeRF and mesh rasterization. Additionally, we do observe a decline in rendering performance when converting NeRF into game engine-compatible meshes.

\subsection{Video2Game}
\label{sect:exp_interaction}
We have shown our approach's effectiveness in rendering quality and reconstruction accuracy across various setups. Next, we demonstrate the construction of a web browser-compatible game enabling player control and interaction with the environment at over 100 FPS.

\noindent \textbf{Data preparation:} 
We build our environments based on videos in Gardenvase~\cite{barron2022mip}, KITTI-360 \cite{liao2022kitti} and VR-NeRF \cite{VRNeRF} mentioned in Sec.~\ref{sec:data}, using our proposed approach. The outcomes include executable environments with mesh geometry, materials, and rigid-body physics, all encoded in GLB and texture files.

\noindent \textbf{Game engine:}
We build our game based on several key components in our game engine mentioned in Sec. \ref{sec:game}.
By leveraging them, our game generates a highly realistic visual rendering as well as physical interactions (see Fig.~\ref{fig:teaser}) and runs smoothly at an interactive rate across various platforms, browsers, and hardware setups (see Tab.~\ref{tab:speed-platform}). 
\hongchi{As for other game engines (see Fig.~\ref{tab:platform}), in Blender~\cite{blender} we showcase the compatibility of our exported assets with other game engines. For Unreal~\cite{unrealengine}, we further demonstrate a real-time game demo where a humanoid robot can freely interact within the Gardenvase scene, such as standing on the table and kicking off the central vase. These prove the compatibility of our proposed pipeline. }

\noindent \textbf{Interactive game features:} 
{\it Movement in games:} Agents can navigate the area freely within the virtual environment where their actions follow real-world physics and are constrained by collision models. 
{\it Shooting game:} 
For realistic shooting physics, we calculated the rigid-body collision dynamics for both the central vase and the surrounding scene (see Fig.~\ref{fig:collision}), separated using mesh semantic filtering. We used a box collider for the vase and convex polygon colliders for the background. The player shoots footballs with a sphere collider at the vase on the table, causing it to fly off and fall to the ground (see Fig.~\ref{fig:teaser}).
{\it Temple-Run like game:} 
The agent collects coins while running in the KITTI Loop composed of four streets in KITTI-360. Obstructive chairs on the road can be smashed thanks to pre-computed fracture animations. The agent can also drive and push roadside vehicles existing in the scene forward by crashing into them. This interactivity is achieved through rigid-body dynamics simulation and collision modeling.

\begin{table}[t]
    \centering
    \vspace{-2mm}
    \resizebox{0.48\textwidth}{!}{
    \begin{tabular}{lcccc}
    \toprule
         & Platform & FPS (hz) & CPU-Usage (\%)  & GPU-Usage (\%) \\ \midrule
       Mac M1 Pro & Mac OS, Chrome & 102 & 34 & 70 \\
       Intel Core i9 + NV 4060 & Windows, Edge & 240 & 6 & 74 \\
       AMD 5950 + NV 3090 & Linux, Chrome & 144 & 20 & 40 \\ \bottomrule
    \end{tabular}
}
    \caption{\textbf{Runtime Analysis.} {Our interactive environment can run in real-time across various hardware setup and various platforms. User actions may slightly vary, which could lead to minor variations in runtime. 
    }}
    \label{tab:speed-platform}
    
\end{table}

\begin{figure}
    \centering
    \includegraphics[width=0.48\textwidth]{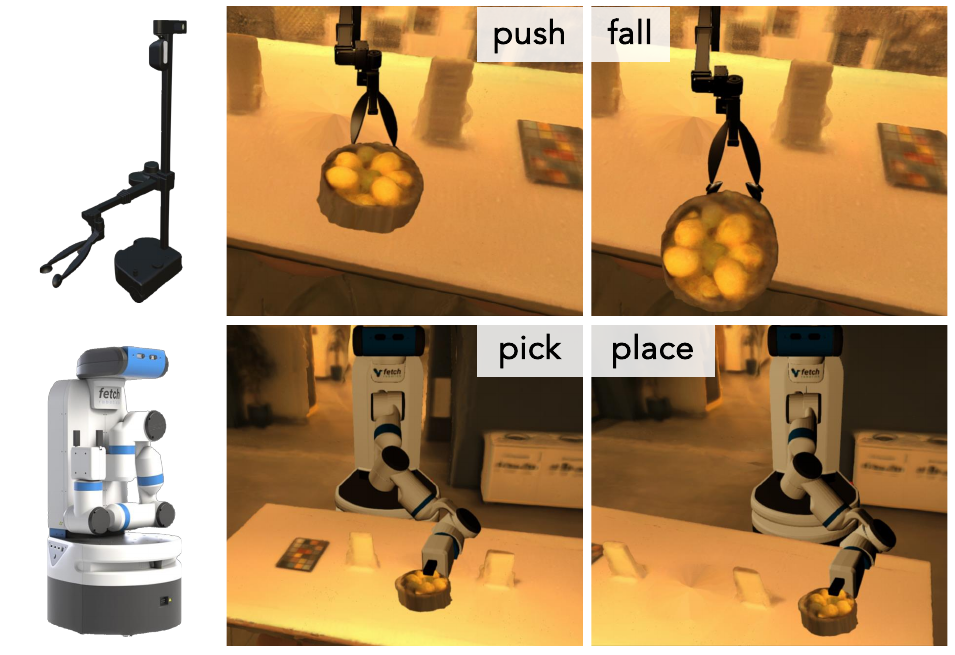}
    \vspace{-5mm}
    \caption{\textbf{Robot simulation in VRNeRF~\cite{VRNeRF} dataset.} We demonstrate the possibility of conducting robot learning in our  virtual environments using Stretch Robot \cite{stretchrb} and Fetch Robot \cite{fetchrb}.}
    \label{fig:vrnerf}
\end{figure}

\noindent \textbf{Robot simulation:} {We demonstrate the potential of leveraging Video2Game for robot simulation using the VRNeRF dataset. We reconstruct the scene and segment simulatable rigid-body objects (\eg, the fruit bowl on the table). We show two demos in Fig. \ref{fig:vrnerf}:  a Stretch Robot pushing the bowl off the table and a Fetch Robot performing pick-and-place actions.
We employ PyBullet~\cite{coumans2021} to simulate the underlying physics with the help of corresponding collision models.
Since real-time grasping simulation is challenging, following existing robot simulation frameworks~\cite{kolve2022ai2thor,Deitke2020RoboTHORAn,Ehsani2021ManipulaTHORA}, 
objects near the Fetch gripper are automatically picked up. 
This demonstrates our model's ability to convert a real-time video stream into a virtual environment, allowing robots to rehearse before acting in the real environment.
}

\section{Conclusion}
We present a novel approach to converting real-world video footage into playable, real-time, and interactive game environments.
Specifically, we combine the potential of NeRF modeling with physics modeling and integrate them into modern game engines. 
Our approach enables any individual to transform their surroundings into an interactive digital environment, unlocking exciting possibilities for 3D content creation, with promising implications for future advancements in digital game design and robot simulation.
\paragraph{Acknowledgements} This project is supported by the Intel AI SRS gift, the IBM IIDAI Grant, the Insper-Illinois Innovation Grant, the NCSA Faculty Fellowship, and NSF Awards \#2331878, \#2340254, and \#2312102. We greatly appreciate the NCSA for providing computing resources. We thank Derek Hoiem and Albert Zhai for helpful discussions. We thank Jingkang Wang, Joao Marques and Hanxiao Jiang for proofreading.

{
    \small
    \bibliographystyle{ieeenat_fullname}
    \bibliography{main}
}
\clearpage
\setcounter{page}{1}
\maketitlesupplementary

\setcounter{section}{0}
\setcounter{footnote}{0}

\appendix
\section{Additional Results and Analysis}
\paragraph{More qualitative results.} We provide more qualitative comparison results among baselines~\cite{muller2022instant,nerfstudio,kerbl20233d,chen2022mobilenerf,yariv2023bakedsdf} and our proposed method. For comparisons between InstantNGP~\cite{muller2022instant}, Nerfacto~\cite{nerfstudio}, 3D Gaussian Splatting~\cite{kerbl20233d} and our base NeRF in KITTI-360 dataset~\cite{liao2022kitti} and Garden scene in Mipnerf-360 Dataset~\cite{barron2022mip}, see Fig.~\ref{tab:supp_qualitative} and Fig.~\ref{tab:supp_qualitative_garden}. We observe that our method renders less noisy geometries while maintaining a superior or comparable visual quality. Especially, 3D Gaussian Splatting~\cite{kerbl20233d} fails to learn correct 3D orientations of Gaussians in sparse settings like KITTI-360~\cite{liao2022kitti}, leading to weird color renderings in novel views and noisy geometry rendering. As for mesh rendering qualitative comparison between \cite{chen2022mobilenerf,yariv2023bakedsdf} and ours, see Fig.~\ref{tab:supp_mesh_qualitative}. Our mesh rendering has similar and comparable rendering results in Garden scene~\cite{barron2022mip}. However, in KITTI-360 dataset~\cite{liao2022kitti} which is extremely large-scale and open, the performance of MobileNeRF~\cite{chen2022mobilenerf} drops dramatically and BakedSDF~\cite{yariv2023bakedsdf} generates slightly blurry in road-aside car rendering, while our mesh rendering is not only superior in KITTI-360 dataset~\cite{liao2022kitti}, but it also maintains stable performance across different datasets. 

\section{Dataset Details}
\subsection{KITTI-360 Dataset}
We build ``KITTI-Loop game'' based on KITTI-360 Dataset~\cite{liao2022kitti}. We use frames from sequence 0. The loop we build in our game utilizes frames 4240-4364, 6354-6577, 7606-7800, and 10919-11050. We compose those four snippets into a closed loop in our game. For baseline comparison and ablation study, we perform experiments on two blocks containing frames 7606-7665 and 10919-11000. We split the validation set every 10 frames (frames 7610, 7620, 7630, 7640, 7650, and 7660 for the first block; frames 10930, 10940, 10950, 10960, 10970, 10980, 10990 for the second block). We report the average metrics of two blocks.
\subsection{Mipnerf-360 Dataset}
We build the ``Gardenvase game'' based on the Garden scene of Mipnerf-360 Dataset~\cite{barron2022mip}. We split the validation set every 20 frames.
\subsection{VRNeRF Dataset}
We build our robot simulation environment based on the ``table'' scene of VRNeRF Dataset~\cite{VRNeRF}.

\section{Video2Game Implementation Details}
\subsection{Base NeRF Training Details}
\begin{figure}[]
    \centering
    \includegraphics[width=\linewidth]{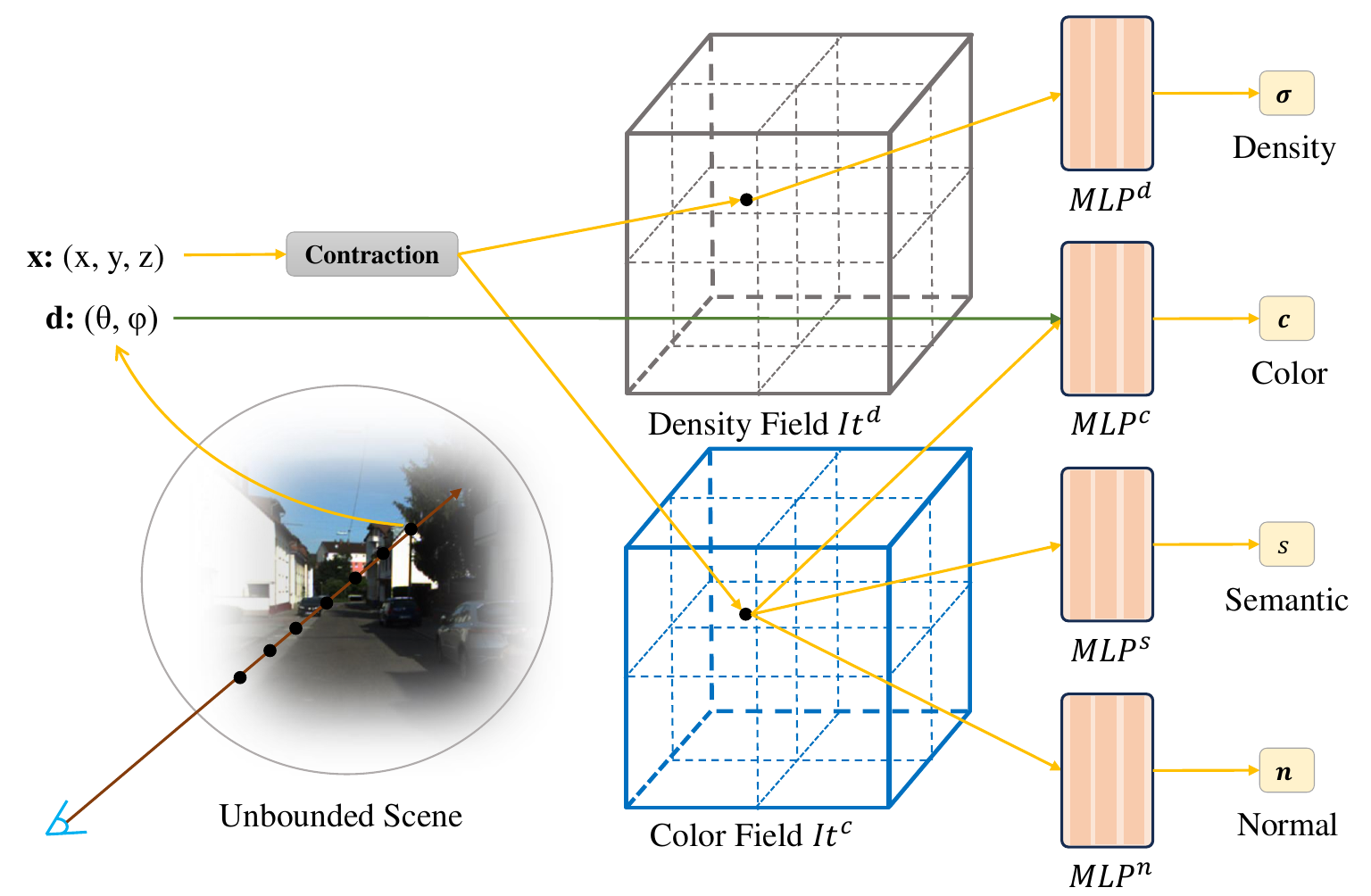}
    \vspace{-6mm}
    \caption{\textbf{Video2Game NeRF Module:}  
    The diagram of our designed NeRF.  
    }
    \label{fig:nerf}
\end{figure}

\paragraph{Network architecture and hyper-parameters} 
Our network consists of two hash grid encoding~\cite{muller2022instant} components $\texttt{It}_{d}$ and $\texttt{It}_c$ and MLP headers $\texttt{MLP}_{\theta_d}^{d}$, $\texttt{MLP}_{\theta_c}^{c}$, $\texttt{MLP}_{\theta_s}^{s}$, and $\texttt{MLP}_{\theta_n}^{n}$, each with two 128 neurons layers inside. Taking 3D position input $\bx$, density $\sigma$ is calculated following $\sigma = \texttt{MLP}_{\theta_d}^{d}(\texttt{It}_{d}(\texttt{Ct}(\bx), \Phi_{d}))$.
Color feature $f=\texttt{It}_{c}(\texttt{Ct}(\bx), \Phi_{c})$. 
Then we calculate $\bc, s, \bn$ from feature $f$ and direction $\bd$ through $\bc = \texttt{MLP}_{\theta_c}^{c}(f, \bd)$, $s = \texttt{MLP}_{\theta_s}^{s}(f)$ and $\bn = \texttt{MLP}_{\theta_n}^{n}(f)$ respectively. All parameters involved in training our base NeRF can be represented as NGP voxel features $\Phi = \{\Phi_d, \Phi_c\}$ and MLP parameters $\theta = \{ {\theta_d}, {\theta_c}, {\theta_s}, {\theta_n} \}$. To sum up, we get $\bc, \sigma, s, \bn = F_\theta(\mathbf{x}, \bd; \Phi) = \texttt{MLP}_\theta(\texttt{It}(\texttt{Ct}(\bx), \Phi), \bd).$ The detailed diagram of our NeRF can be found in Fig.~\ref{fig:nerf}.

Our hash grid encoding~\cite{muller2022instant} is implemented by tiny-cuda-nn~\cite{tiny-cuda-nn}, and we set the number of levels to 16, the dimensionality of the feature vector to 8 and Base-2 logarithm of the number of elements in each backing hashtable is 19 for $\texttt{It}_{d}$ and 21 for $\texttt{It}_c$. As for activation functions, we use ReLU~\cite{nair2010rectified} inside all MLPs,  Softplus for density $\sigma$ output, Sigmoid for color $\bc$ output, Softmax for semantic $s$ output and no activation function for normal $\bn$ output (We directly normalize it instead).
\paragraph{KITTI-Loop additional training details}
In KITTI-Loop which uses KITTI-360 Dataset~\cite{liao2022kitti}, we also leverage stereo depth generated from DeepPruner~\cite{Duggal2019ICCV}. Here we calculate the actual depth from disparity and the distance between binocular cameras and adopt L1 loss to regress. We \textbf{haven't} used any LiDAR information to train our base NeRF in KITTI-360 Dataset~\cite{liao2022kitti}.
\subsection{Mesh Extraction and Post-processing Details}
\paragraph{Mesh Post-processing details} In mesh post-processing, we first utilize all training camera views to prune the vertices and faces that can't be seen. Next, we delete those unconnected mesh components that have a small number of faces below a threshold so as to delete those floaters in the mesh. Finally, we merge close vertices in the mesh, then perform re-meshing using PyMesh~\cite{pymesh} package, which iteratively splits long edges over a threshold, merges short edges below a threshold and removes obtuse triangles. Remeshing helps us get better UV mapping results since it makes the mesh ``slimmer" (less number of vertices and faces) and has similar lengths of edges. After the post-processing, we get meshes with a relatively small number of vertices and faces while still effectively representing the scene.  
\paragraph{Special settings in KITTI-Loop}
In KITTI-Loop, we partition the whole loop into 14 overlapping blocks. Since we adopt pose normalization and contract space in each block when training, it needs alignments when we compose them together. For each block, we first generate its own mesh. We partition the whole contract space ($[-1, 1]^3$) into 3*3*3 regions, and perform marching cubes with the resolution of 256*256*256 in each region. We then transform those vertices back from contract space to the coordinates before contraction. We then perform mesh post-processing here. To compose each part of the mesh in KITTI-Loop together, we then transform the mesh to KITTI-Loop world coordinates. For those overlapping regions, we manually define the block boundary and split the mesh accordingly. Finally, we add a global sky dome over the KITTI-Loop.

\begin{table*}[t]
    \centering
    \setlength\tabcolsep{0.05em} %
    \resizebox{\textwidth}{!}{
    \begin{tabular}{lccc}
           & RGB & Depth & Normal \\ 
          \raisebox{7mm}[0pt][0pt]{\rotatebox[origin=c]{90}{G.T.}} & 
          \includegraphics[width=0.33\textwidth]{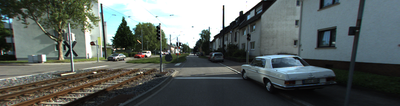} & 
          \includegraphics[width=0.33\textwidth]{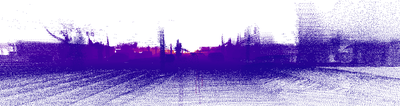}  & 
          \includegraphics[width=0.33\textwidth]{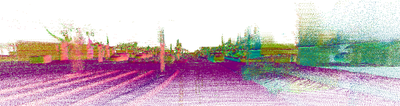} \\
         \raisebox{7mm}[0pt][0pt]{\rotatebox[origin=c]{90}{InstantNGP}} & 
         \includegraphics[width=0.33\textwidth]{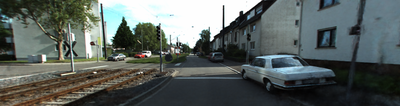} & 
          \includegraphics[width=0.33\textwidth]{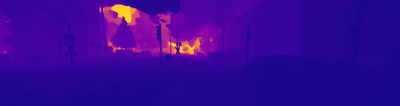}  & 
          \includegraphics[width=0.33\textwidth]{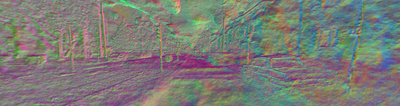} \\
          \raisebox{7mm}[0pt][0pt]{\rotatebox[origin=c]{90}{Nerfacto}} & 
          \includegraphics[width=0.33\textwidth]{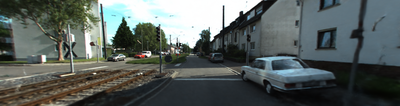} & 
          \includegraphics[width=0.33\textwidth]{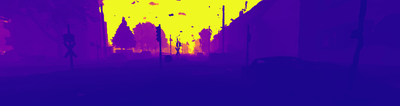}  & 
          \includegraphics[width=0.33\textwidth]{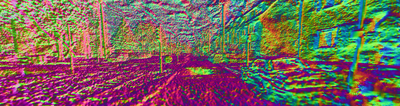} \\
         \raisebox{7mm}[0pt][0pt]{\rotatebox[origin=c]{90}{Guass. Spl.}}& 
         \includegraphics[width=0.33\textwidth]{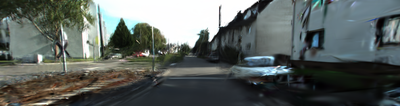} & 
         \includegraphics[width=0.33\textwidth]{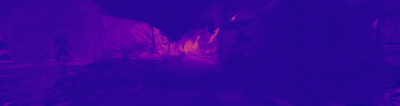} & 
          \raisebox{7mm}[0pt][0pt]{N/A} \\
         \raisebox{7mm}[0pt][0pt]{\rotatebox[origin=c]{90}{Ours}} & 
         \includegraphics[width=0.33\textwidth]{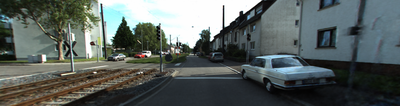} & 
          \includegraphics[width=0.33\textwidth]{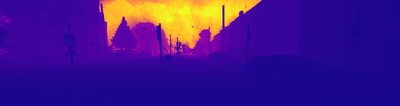}  & 
          \includegraphics[width=0.33\textwidth]{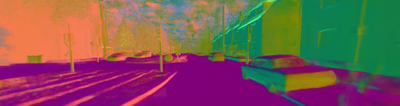} \\

    \end{tabular}
    }
    \captionof{figure}{{\bf Qualitative comparisons among NeRF models~\cite{muller2022instant,nerfstudio} and 3D Gaussian Splatting~\cite{kerbl20233d} in KITTI-360 Dataset~\cite{liao2022kitti}.} We provide NeRF rendering depths and normals for comparison as well. For 3D Gaussian Splatting, only rendering depth is provided. Here we consider depths measured by LiDAR point cloud in KITTI-360 and compute normals based on it as our ground truth.
    }
    \label{tab:supp_qualitative}
\end{table*}

\begin{table*}[t]
    \centering
    \setlength\tabcolsep{0.05em} %
    \resizebox{\textwidth}{!}{
    \begin{tabular}{lccc}
           & RGB & Depth & Normal \\ 
          \raisebox{18mm}[0pt][0pt]{\rotatebox[origin=c]{90}{G.T.}} & 
          \includegraphics[width=0.33\textwidth]{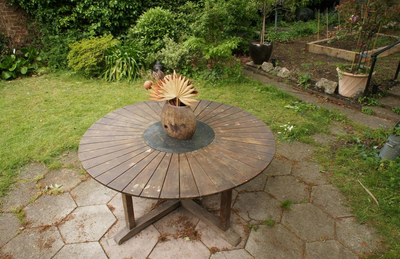} & 
          \raisebox{18mm}[0pt][0pt]{N/A} &
          \raisebox{18mm}[0pt][0pt]{N/A} \\
          \raisebox{18mm}[0pt][0pt]{\rotatebox[origin=c]{90}{InstantNGP}} & 
          \includegraphics[width=0.33\textwidth]{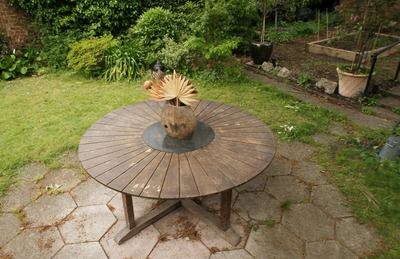} & 
          \includegraphics[width=0.33\textwidth]{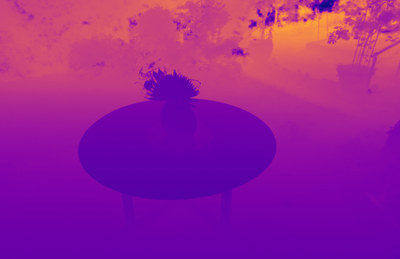} &
          \includegraphics[width=0.33\textwidth]{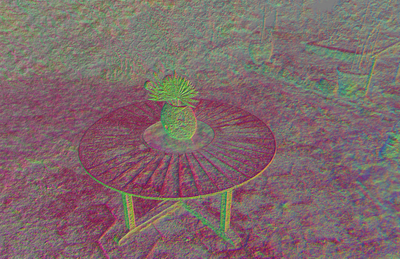} \\
          \raisebox{18mm}[0pt][0pt]{\rotatebox[origin=c]{90}{Nerfacto}} & 
          \includegraphics[width=0.33\textwidth]{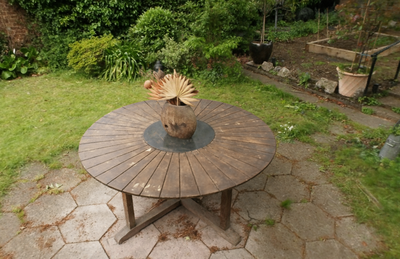} & 
          \includegraphics[width=0.33\textwidth]{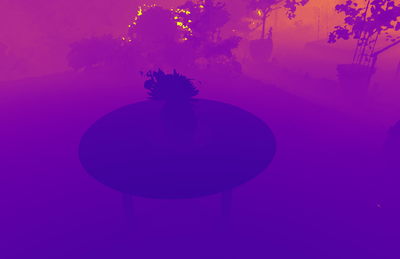} &
          \includegraphics[width=0.33\textwidth]{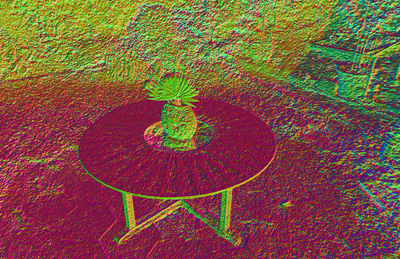} \\
          \raisebox{18mm}[0pt][0pt]{\rotatebox[origin=c]{90}{Gauss. Spl.}} & 
          \includegraphics[width=0.33\textwidth]{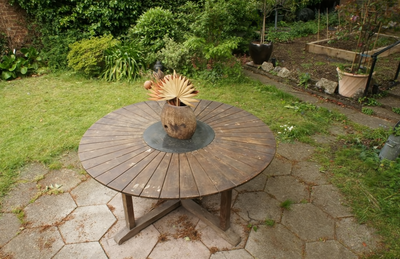} & 
          \includegraphics[width=0.33\textwidth]{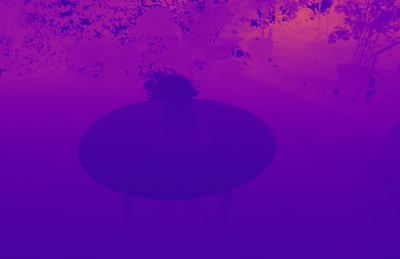} &
          \raisebox{18mm}[0pt][0pt]{N/A} \\
        \raisebox{18mm}[0pt][0pt]{\rotatebox[origin=c]{90}{Ours}} & 
          \includegraphics[width=0.33\textwidth]{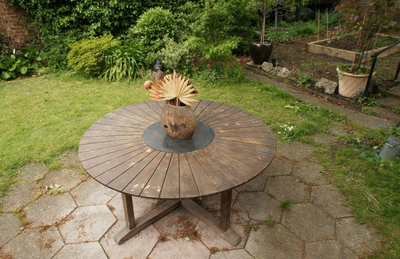} & 
          \includegraphics[width=0.33\textwidth]{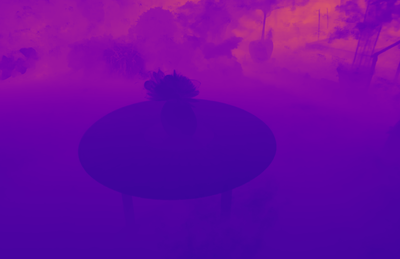} &
          \includegraphics[width=0.33\textwidth]{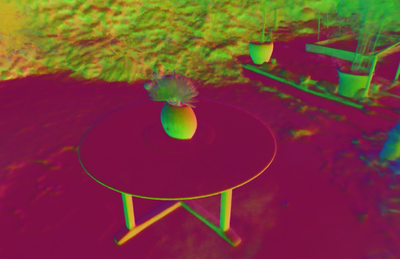} \\

    \end{tabular}
    }
    \captionof{figure}{{\bf Qualitative comparisons among NeRF models~\cite{muller2022instant,nerfstudio} and 3D Gaussian Splatting~\cite{kerbl20233d} in Garden scene~\cite{barron2022mip}.} We provide NeRF rendering depths and normals for comparison as well. For 3D Gaussian Splatting, only rendering depth is provided.
    }
    \label{tab:supp_qualitative_garden}
\end{table*}

\begin{table*}[t]
    \centering
    \setlength\tabcolsep{0.05em} %
    \resizebox{\textwidth}{!}{
    \begin{tabular}{cccc}
        \includegraphics[width=0.24\textwidth]{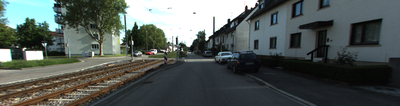} & 
        \includegraphics[width=0.24\textwidth]{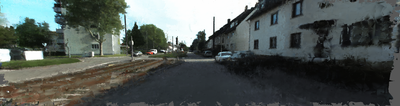} & 
        \includegraphics[width=0.24\textwidth]{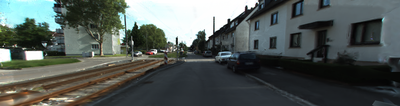} & 
        \includegraphics[width=0.24\textwidth]{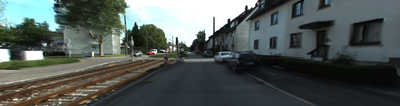} \\
        
        \includegraphics[width=0.24\textwidth]{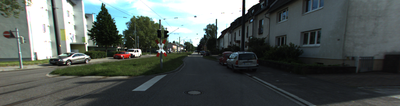} & 
        \includegraphics[width=0.24\textwidth]{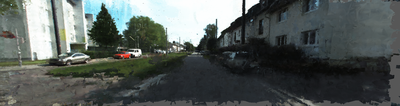} & 
        \includegraphics[width=0.24\textwidth]{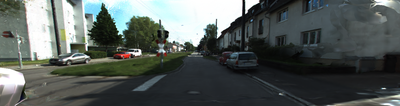} & 
        \includegraphics[width=0.24\textwidth]{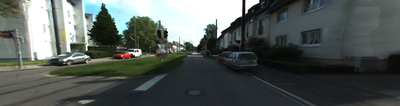} \\
        
        \includegraphics[width=0.24\textwidth]{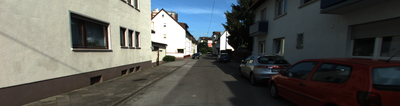} & 
        \includegraphics[width=0.24\textwidth]{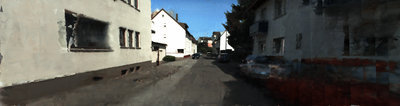} & 
        \includegraphics[width=0.24\textwidth]{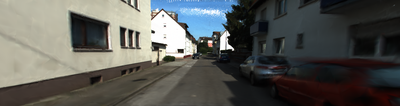} & 
        \includegraphics[width=0.24\textwidth]{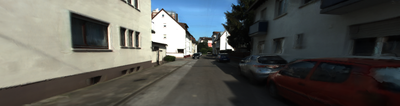} \\
        
        \includegraphics[width=0.24\textwidth]{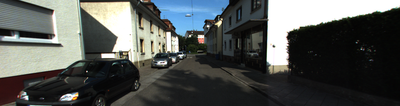} & 
        \includegraphics[width=0.24\textwidth]{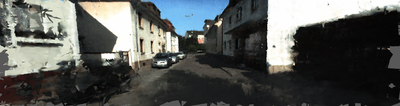} & 
        \includegraphics[width=0.24\textwidth]{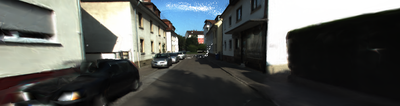} & 
        \includegraphics[width=0.24\textwidth]{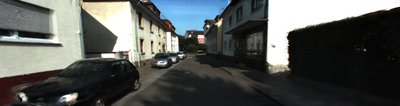} \\

         G.T. & MobileNeRF & BakedSDF & Ours
       
    \end{tabular}
    }
    \captionof{figure}{{\bf Qualitative comparisons in mesh rendering.} We compare our proposed mesh rendering method to MobileNeRF~\cite{chen2022mobilenerf} and BakedSDF~\cite{yariv2023bakedsdf} in KITTI-360 Dataset~\cite{liao2022kitti}.
    }
    \label{tab:supp_mesh_qualitative}
\end{table*}

\subsection{NeRF Baking Details}
For each extracted mesh, we bake the NeRF's color and specular components to it with nvdiffrast~\cite{Laine2020diffrast}. 
\paragraph{GLSL MLP settings} We adopt a two-layer tiny MLP with 32 hidden neurons. We use ReLU~\cite{nair2010rectified} activation for the first layer and sigmoid for the second. We re-implement that MLP with GLSL code in Three.js renderer's shader.
\paragraph{Initialization of texture maps and MLP shader} 
Training the textures $\bT = [\bB; \bS]$ and MLP shader $\texttt{MLP}_\theta^\text{shader}$ all from scratch is slow. 
Instead, we adopt an initialization procedure. 
Inspired by \cite{tang2022nerf2mesh, yariv2023bakedsdf}, we encode the 3D space by hash encoding~\cite{muller2022instant} $\texttt{It}^{M}$ and an additional MLP $\texttt{MLP}_{\theta_0}^{M}$.
Specifically, we first rasterize the mesh into screen space, obtain the corresponding 3D position $x_i$ on the surface of the mesh within each pixel, transform it into contract space $\texttt{Ct}(x_i)$, and then feed it into $\texttt{It}^{M}$ and $\texttt{MLP}_{\theta_0}^{M}$ to get the base color $\mathbf{B}_i$ and specular feature $\mathbf{S}_i$, represented as $\mathbf{B}_i$, $\mathbf{S}_i = \texttt{MLP}_{\theta_0}^{M}(\texttt{It}^{M}(\texttt{Ct}(x_i),  \Phi_0))$.
Finally we computes the sum of the view-independent base color $\mathbf{B}_i$ and the view-dependent specular color following $\bC_\text{R} = \mathbf{B}_i + \texttt{MLP}_\theta^\text{shader}(\mathbf{S}_i, \bd_i)$.
The parameters ${\Phi_0, \theta_0, \theta}$ are optimized by minimizing the color difference between the mesh model and the ground truth: $\mathcal{L}^\text{render}_{\text{initialize} {\Phi_0, \theta_0, \theta}} = \sum_{\br}  \| \bC_\text{R}(\br) - \bC_\text{GT}(\br) \|_2^2$. 
Anti-aliasing is also adopted in the initialization step by perturbing the optical center of the camera.
With learned parameters, every corresponding 3D positions $x_i$ in each pixel of 2D unwrapped texture maps $\bT = [\bB; \bS]$ is initialized following $\mathbf{B}_i$, $\mathbf{S}_i = \texttt{MLP}_{\theta_0}^{M}(\texttt{It}^{M}(\texttt{Ct}(x_i),  \Phi_0))$ and the parameters of $\texttt{MLP}_\theta^\text{shader}$ is directly copied from initialization stage. 

\subsection{Physical Module Details}
\paragraph{Physical dynamics} It is important to note that our approach to generating collision geometries is characterized by meticulous design. In the case of box collider generation, we seamlessly repurpose the collider used in scene decomposition. When it comes to triangle mesh colliders, we prioritize collision detection efficiency by simplifying the original mesh. Additionally, for convex polygon colliders, we leverage V-HACD~\cite{vhacd} to execute a precise convex decomposition of the meshes.

\hongchi{
\paragraph{Physical parameters assignments.} 
Physical parameters for static objects, such as the ground, were set to default values. For interactive instances like cars and vases, we could query GPT-4 with box highlights and the prompts as shown on the left. Note that we reason about mass and friction using the same prompt. The output is a range, and we find that selecting a value within this range provides reasonable results. See Fig.~\ref{fig:gpt4} for an example.
Unit conversion from the metric system to each engine's specific system is needed.
}

\begin{table}[t]
    \centering
    \setlength\tabcolsep{0.05em} %
    \resizebox{0.48\textwidth}{!}{
    \begin{tabular}{c}
          \includegraphics{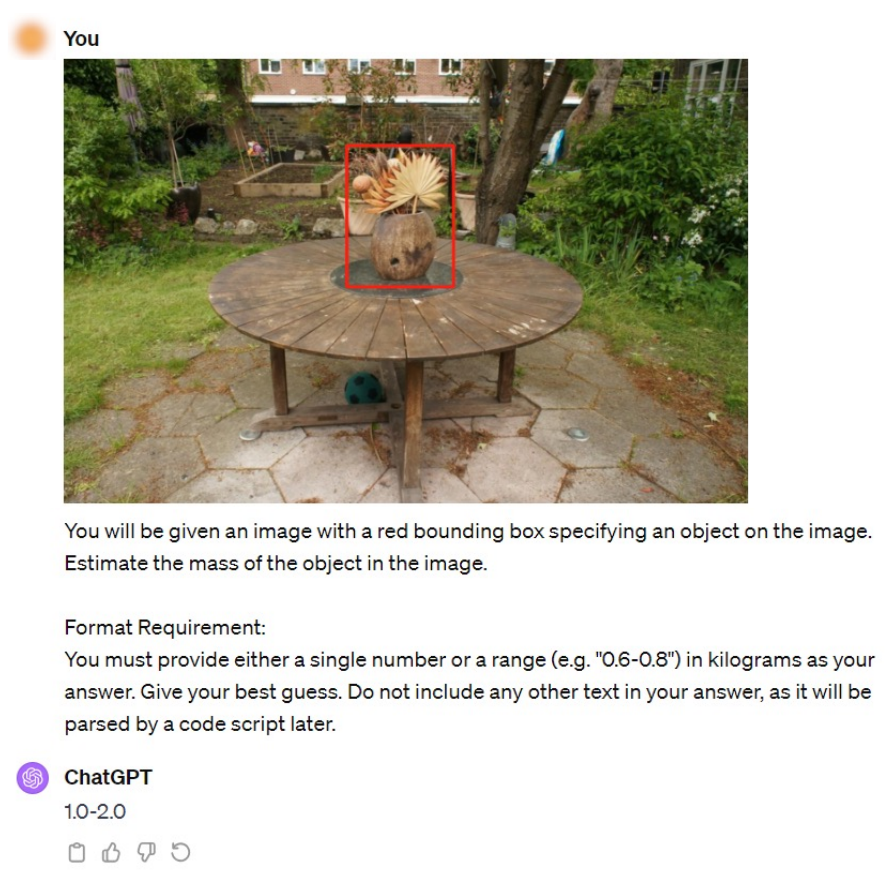}\\ 
    \end{tabular}
    }
    \vspace{-2mm}
    \captionof{figure}{{\bf Example of physical property reasoning by GPT-4.}}
    \label{fig:gpt4}
\end{table}

\subsection{Robot Simulation Details}
\paragraph{Data preparation} 
We demonstrate the potential of leveraging Video2Game for robot simulation using the VRNeRF~\cite{VRNeRF} dataset. We reconstruct the scene and segment simulatable rigid-body objects (\eg, the fruit bowl on the table). Then collision models are generated for those physical entities for subsequent physical simulations.
\paragraph{Physical simulation} 
To simulate the interactions between robots and physical entities in a dynamic environment, we employ PyBullet~\cite{coumans2021}, a Python module designed for physics simulations in the realms of games, robotics, and machine learning. Given the intricate dynamics of articulated robots, PyBullet serves as a powerful tool for conducting physics calculations within the context of robot simulation. Our approach involves loading all generated collision models and URDF~\footnote{http://wiki.ros.org/urdf} files for both the Stretch Robot \cite{stretchrb} and Fetch Robot \cite{fetchrb}.
Utilizing PyBullet's integrated robotic inverse kinematics, we can effectively control the mechanical arms of the robots to interact with surrounding objects. Specifically, for the Stretch Robot, we define a predefined path for its arm, enabling it to exert a direct force to displace the central bowl off the table. On the other hand, for the Fetch Robot, we leverage the collision boxes specified in its URDF file. Our manipulation involves grasping the corresponding collision model of the central bowl on the table, eschewing the use of the magnetic gripper for object control. Subsequently, the robot lifts the bowl and relocates it to a different position.
Following the simulations in PyBullet, we extract physics calculation results, including joint values and the position of the robots' base link. These results are then exported and integrated into the rendering engine of Three.js for further visualization and analysis.
\paragraph{Rendering in robot simulation} 
We import the URDF files of our robots into our engine using the urdf-loader \cite{urdfloader} in Three.js, a library that facilitates the rendering and configuration of joint values for the robots. Leveraging pre-computed physics simulations in PyBullet, which are based on our collision models, we seamlessly integrate these simulations into the Three.js environment. This integration allows us to generate and render realistic robot simulation videos corresponding to the simulated physics interactions.

\subsection{Training time}
For base NeRF training, it takes 8 hours for training 150k iterations on an A6000. For the NeRF baking procedure, the initialization and training take 4 hours on an A5000.

\section{Baseline Details}
\subsection{Instant-NGP}
We adopt the re-implementation of Instant-NGP~\cite{muller2022instant} in \cite{ngpimpl}. We choose the best hyper-parameters for comparison. For normal rendering, we calculate by the derivative of density value.
\subsection{Nerfacto}
Nerfacto is proposed in Nerfstudio~\cite{nerfstudio}, an integrated system of simplified end-to-end process of creating, training, and testing NeRFs. We choose their recommended and default settings for the Nerfacto method.
\subsection{3D Gaussian Splatting}
For 3D Gaussian Splatting~\cite{kerbl20233d} training in Garden scene~\cite{barron2022mip}, we follow all their default settings. In the KITTI-360 Dataset, there are no existing SfM~\cite{schonberger2016structure} points. We choose to attain those 3D points by  LoFTR~\cite{sun2021loftr} 2D image matching and triangulation in Kornia~\cite{eriba2019kornia} using existing camera projection matrixs and matching results. We choose their best validation result throughout the training stage by testing every 1000 training iterations.
\subsection{MobileNeRF}
In Garden scene~\cite{barron2022mip}, we directly follow the default settings of MobileNeRF~\cite{chen2022mobilenerf}. For training in KITTI-360 Dataset~\cite{liao2022kitti}, we adopt their ``\textit{unbounded} 360 scenes'' setting for the configurations of polygonal meshes, which is aligned with KITTI-360 Dataset. 
\subsection{BakedSDF}
We adopt the training codes of BakedSDF~\cite{yariv2023bakedsdf} in SDFStudio~\cite{Yu2022SDFStudio}, from which we can attain the exported meshes with the resolution of 1024x1024x1024 by marching cubes. For the baking stage, we adopt three Spherical Gaussians for every vertices and the same hyper-parameters of NGP~\cite{muller2022instant} mentioned in \cite{yariv2023bakedsdf}. We follow the notation BakedSDF~\cite{yariv2023bakedsdf} used in its paper, where ``offline'' means volume rendering results.

\section{Limitation}
\hongchi{Although Video2Game framework could learn view-dependent visual appearance through its NeRF module and mesh module, it doesn't learn necessary material properties for physics-informed relighting, such as the metallic property of textures. Creating an unbounded, \emph{relightable} scene from a single video, while extremely challenging, can further enhance realism. We leave this for future work.}

\end{document}